\documentclass[conference]{IEEEtran}
\IEEEoverridecommandlockouts
\usepackage{cite}

\usepackage{amsmath,amssymb,amsfonts}
\usepackage{algorithmic}
\usepackage{graphicx}
\usepackage{textcomp}
\usepackage{xcolor}
\usepackage[table]{xcolor} 
\usepackage{adjustbox}
\usepackage{stfloats}
\usepackage{booktabs}
\usepackage{float}
\usepackage{stfloats}
\usepackage{url}
\usepackage[most]{tcolorbox}

\tcbset{
  myprompt/.style={
    enhanced,
    breakable,
    listing only,
    listing options={
      basicstyle=\ttfamily\small,
      breaklines=true,
      columns=fullflexible,
      keepspaces=true,
      showstringspaces=false,
      framexleftmargin=0pt,
      framexrightmargin=0pt,
      xleftmargin=0pt,
      xrightmargin=0pt
    },
    colback=white,
    colframe=black,
    boxrule=0.6pt,
    left=3pt, right=3pt, top=3pt, bottom=3pt, boxsep=2pt
  }
}

\usepackage[colorlinks=true, allcolors=blue, citecolor=blue]{hyperref}

\def\BibTeX{{\rm B\kern-.05em{\sc i\kern-.025em b}\kern-.08em
    T\kern-.1667em\lower.7ex\hbox{E}\kern-.125emX}}
\begin{document}

\title{Assessing Large Language Models for Medical QA: Zero-Shot and LLM-as-a-Judge Evaluation}

\author{
    \IEEEauthorblockN{Shefayat E Shams Adib, Ahmed Alfey Sani, Ekramul Alam Esham, Ajwad Abrar, Tareque Mohmud Chowdhury\\}

    \IEEEauthorblockA{Department of Computer Science and Engineering, Islamic University of Technology, Gazipur, Bangladesh\\}
    
    \IEEEauthorblockA{Email: \{shefayatadib, ahmedalfey, ekramulalam, ajwadabrar, tareque\}@iut-dhaka.edu}
}

\makeatletter
\let\old@ps@IEEEtitlepagestyle\ps@IEEEtitlepagestyle
\def\confheader#1{%
    \def\ps@IEEEtitlepagestyle{%
        \old@ps@IEEEtitlepagestyle%
        \def\@oddhead{\strut\hfill#1\hfill\strut}%
        \def\@evenhead{\strut\hfill#1\hfill\strut}%
    }%
    \ps@headings%
}
\makeatother

\confheader{
        \parbox{20cm}{2025 28th International Conference on Computer and Information Technology (ICCIT)\\19-21 December 2025, Cox’s Bazar, Bangladesh}
}

\IEEEpubid{
\begin{minipage}[t]{\textwidth}\ \\[10pt]
      \small{Accepted in 28th ICCIT, 2025 }  
\end{minipage}
}

\maketitle

\begin{abstract}
Recently, Large Language Models (LLMs) have gained significant traction in medical domain, especially in developing a QA systems to Medical QA systems for enhancing access to healthcare in low-resourced settings. This paper compares five LLMs deployed between April 2024 and August 2025 for medical QA, using the iCliniq dataset, containing 38,000 medical questions and answers of diverse specialties. Our models include Llama-3-8B-Instruct, Llama 3.2 3B, Llama 3.3 70B Instruct, Llama-4-Maverick-17B-128E-Instruct, and GPT-5-mini. We are using a zero-shot evaluation methodology and using BLEU and ROUGE metrics to evaluate performance without specialized fine-tuning. Our results show that larger models like Llama 3.3 70B Instruct outperform smaller models, consistent with observed scaling benefits in clinical tasks. It is notable that, Llama-4-Maverick-17B exhibited more competitive results, thus highlighting evasion efficiency trade-offs relevant for practical deployment. These findings align with advancements in LLM capabilities toward professional-level medical reasoning and reflect the increasing feasibility of LLM-supported QA systems in the real clinical environments. This benchmark aims to serve as a standardized setting for future study to minimize model size, computational resources and to maximize clinical utility in medical NLP applications. 
\end{abstract}

\begin{IEEEkeywords}
Large Language Models, Medical Question Answering, Zero-shot Evaluation, Healthcare NLP, LLM-as-a-Judge
\end{IEEEkeywords}

\section{Introduction}
Recent advances in Large Language Models (LLMs) have demonstrated promising capabilities in medical research, showing enhanced decision-making and assessment abilities within healthcare domains \cite{singhal2025toward, chen2025benchmarking}. While achieving strong performance across various biomedical tasks \cite{abrar2025faithful}, LLMs have faced criticism for displaying unfairness in broader NLP applications \cite{abrar2025religious}. Rigorous testing to ensure accuracy, safety, and clinical utility becomes essential before widespread healthcare adoption of these technologies \cite{shool2025systematic, bedi2024systematic}.  

Current evaluation practices attempt to capture healthcare realities by leveraging datasets derived from multiple-choice medical examinations, which provide strong benchmarks for assessing precision \cite{chen2025benchmarking, wu2025bridge}. Notably, DeepSeek-R1 has performed on par with, if not better, than the proprietary models on such benchmarks, while also revealing considerable performance variation across different model architectures \cite{wu2025bridge}.  

Medical applications continue to present considerable challenges for LLMs despite technological advances. A persistent gap between benchmark performance and real-world clinical effectiveness has been highlighted in recent studies \cite{shool2025systematic, bedi2024systematic}. Current models exhibit competence in medical question answering; however, they frequently lack the specialized medical knowledge depth required for complex decision-making \cite{chen2025benchmarking, shool2025systematic}. Given the accelerating pace of LLM development and increasing demand for AI-assisted healthcare in resource-limited contexts, comprehensive evaluation of contemporary model generations becomes essential for determining their practical applicability in medical settings.

We selected five state-of-the-art LLMs with diverse architectural approaches and parameter scales to address this critical gap: Llama-3-8B-Instruct, Llama 3.2 3B, Llama 3.3 70B Instruct, Llama-4-Maverick-17B-128E-Instruct, and GPT-5-mini. Released between April 18, 2024, and August 7, 2025, these models range from efficiency-focused 3B parameter variants to large-scale 70B parameter architectures, featuring various innovations including mixture-of-experts design and enhanced instruction-following capabilities \cite{dubey2024llama, dubey2024llama}. For our evaluation, we utilized the iCliniq \footnote{\url{https://www.kaggle.com/datasets/henry41148/icliniq-medical-qa-38k}} dataset, which contains well-structured medical questions and answers covering diverse medical specialties. The dataset provides accurate clinical contexts across a wide range of medical domains. This comprehensive coverage enables proper assessment of LLMs across various medical knowledge and reasoning tasks \cite{abrantes2025assessing}. 

This evaluation aims to compare the performance of the five mentioned well-capable LLMs across the iCliniq dataset, by applying zero-shot evaluation methodology and computing 
\begin{figure*}[!hbt]
    \centering
    \includegraphics[width=0.9\textwidth]{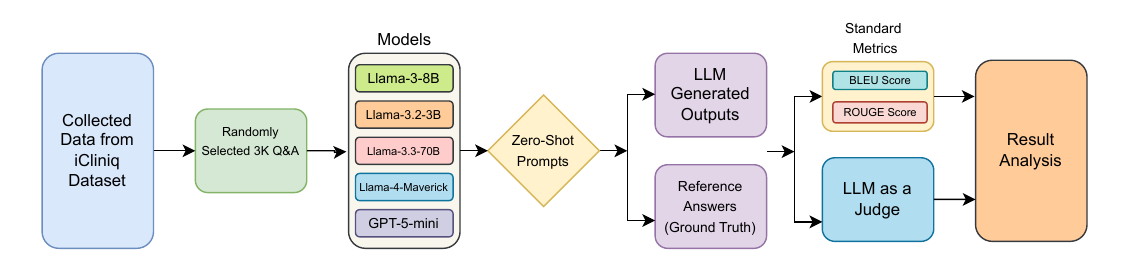} 
    \caption{Methodology pipeline for evaluating LLM performance on medical question answering tasks.}
    \label{fig:placeholder}
\end{figure*}
the BLEU and ROUGE scores to assess the performance of each of them \cite{chen2025benchmarking}. This will help us understand the current capabilities and limitations of LLMs in the healthcare sector so that research for development in AI-assisted healthcare applications becomes more reachable. The study's key contributions are as follows:

\begin{itemize}
    \item \textbf{Comprehensive Multi-Dimensional Benchmarking:} We provide zero-shot evaluation of five advanced LLMs (Llama-3-8B-Instruct, Llama 3.2 3B, Llama 3.3 70B Instruct, Llama-4-Maverick-17B-128E-Instruct, GPT-5-mini) on the iCliniq medical QA dataset using traditional automatic metrics (BLEU, ROUGE) and LLM-as-a-Judge framework evaluating medical accuracy, completeness, safety, clarity, and helpfulness.

    \item \textbf{Parameter Scaling and Clinical Deployment Analysis:} We demonstrate correlations between model size and medical QA performance, revealing Llama 3.3 70B's superior performance and Llama-4-Maverick 17B's competitive efficiency with fewer parameters. The analysis provides actionable deployment guidance for high-accuracy clinical environments versus resource-constrained settings.

    \item \textbf{Standardized Dual-Evaluation Framework:} A methodology combining automatic metrics with LLM-as-a-Judge clinical quality assessment using five specialized criteria addresses n-gram limitations and establishes benchmarking protocols for future medical LLM research.
\end{itemize}


\section{Literature Review}
The rapid progress of Large Language Models (LLMs) has driven expanding adoption in the biomedical healthcare domain, particularly for medical question answering (QA). Early developments focused on domain-specific pretraining, with models like BioBERT and ClinicalBERT demonstrating improved performance on clinical notes and structured data. Med-BERT further expanded this approach to electronic health records \cite{narula2023testing}. However, when applied to generative QA tasks where precision and security are paramount, these early models remained constrained in their capabilities.

The establishment of more complex models marked a defining shift in medical LLM development. Med-PaLM 2 achieved expert-level clinical reasoning with 86.5\% accuracy on the MedQA dataset, demonstrating unprecedented success on medical license exams \cite{singhal2025toward}. The introduction of general-purpose models like GPT-3 and GPT-4 brought notable advancements, with Med-PaLM and Med-PaLM-2 achieving near-expert performance on medical licensing tests and benchmarks \cite{singhal2025toward}. Domain-specific applications further demonstrated efficiency gains through targeted customization, such as VaxBot-HPV, which achieved excellent accuracy and fidelity scores for vaccine-related queries while reducing hallucinations through specialized fine-tuning \cite{li2025vaxbot}.

Research evolution has continued with reasoning-enhanced models and collaborative frameworks across multiple LLMs, showing promise in reducing inter-model disagreement and increasing diagnostic precision \cite{shang2025collaboration}. However, subsequent research revealed persistent challenges: hallucinations and inconsistent reasoning remained problematic in real-world clinical datasets despite strong benchmark performance \cite{lucas2024reasoning, bedi2024systematic}. MedLM research highlighted that while prompt engineering and fine-tuning enhanced output quality, they could not completely resolve reliability and factual grounding issues \cite{narula2023testing}.

Recent research has emphasized broader, more comprehensive evaluations. Ethics and reasoning-focused datasets like MedEthicsQA and MEDIQ have exposed critical flaws in sensitive clinical decision-making scenarios \cite{wei2025medethicsqa, li2024mediq}. Benchmarking efforts have revealed significant performance gaps across model sizes and architectures \cite{chen2025benchmarking, shool2025systematic}, with performance varying considerably across languages and contexts, as demonstrated by multilingual benchmarks such as BRIDGE \cite{wu2025bridge}. Parallel investigations have explored domain-targeted fine-tuning approaches and collaborative methods combining multiple models to minimize inconsistency \cite{shang2025collaboration}. Knowledge-augmented techniques like KoSEL have aimed to improve factual accuracy, though often at the cost of computational efficiency \cite{zeng2025kosel}.

Current research indicates that larger models, such as Llama 3.3 70B, consistently outperform smaller variants \cite{dubey2024llama}. However, the overall advancement in biomedical QA has been less dramatic than anticipated. Despite steadily rising benchmark scores, fundamental challenges persist: hallucinations, inconsistent ethical reasoning, and limited applicability to real patient inquiries remain problematic. This gap emphasizes the necessity of thorough, dataset-driven assessments using real-world medical data, such as those conducted with iCliniq, to evaluate genuine performance improvements rather than relying solely on benchmark achievements. Our study directly addresses this critical gap by benchmarking recent models, including Llama 3.x series, Llama-4 Maverick, and GPT-5-mini, to assess whether newer model generations deliver measurable enhancements in practical medical QA applications.

\section{Methodology}

This paper presents a thorough zero-shot evaluation framework to evaluate five state-of-the-art large language models' performance on answering medical questions. Our adopted methodology adheres to established standards in medical NLP assessment \cite{chen2025benchmarking, shool2025systematic} while modifying the unique requirements of zero-shot situations.

\subsection{Dataset}
We have used the iCliniq Medical QA dataset, an extensive collection of 38,000 real-world medical questions and expert-validated responses from a variety of medical specialties. The dataset, covering a wide range of  medical illnesses, treatments and other health concerns, represents authentic patient inquiries made to medical practitioners. From this large dataset, to ensure computational feasibility, we randomly selected 3,000 question-answer pairs for our evaluation while retaining statistical significance and representativeness across medical fields.

The chosen subset preserves the original distribution of medical specialties included in the entire dataset, such as general medicine, cardiology, dermatology, gastroenterology and  pediatrics. Each question-answer pair in our evaluation set consists of a natural language patient query and an expert-validated response. Given the complexity and variability of medical consultations, the average question length is approximately 126 tokens,  while reference answers average length is 192 tokens\footnote{\url{https://www.kaggle.com/code/ahmedalfeysani/calculating-the-tokens}}.

\begin{table*}[t]  
\centering
\caption{Sample Question-Answer Pair from the iCliniq Dataset}
\label{table:sample_qa}
\footnotesize
\renewcommand{\arraystretch}{1.2}
\setlength{\tabcolsep}{4pt}
\begin{tabular}{p{0.95\linewidth}}
\toprule
\textbf{Question:} Hello doctor, I have recently been diagnosed with type 2 diabetes. Could you provide me with a simplified explanation of how this condition affects my body and what steps I can take to keep my blood sugar levels stable throughout the day? \\
\midrule
\textbf{Answer:} Hello, Welcome to icliniq.com. I read your query and understand your concern. If you have already been diagnosed with diabetes, it is mandatory to keep it under control at all times with the help of diet. Exercise and maintain an active lifestyle and medications prescribed by the treating doctor. Walking, yoga, and exercise as health permits started gradually will help. Check your blood sugar levels and other parameters like liver function test, kidney function test, and lipid profile regularly. Watch what you eat. Small frequented meals might help. Stay away from sweets, sugar, honey, jaggery, marmalade, coconut water, juices, biscuits, rusk, and junk packaged food. Avoid potatoes, white rice, green peas, beetroot, and carrot. Take the dieticians opinion for a detailed diet chart depending on height weight religion culture body mass index likes and dislikes. Avoid stress and tension. Deep breathing and meditation will help. Go for green leafy vegetables, salad, and fruits take only papaya, guava, and kiwi. Replace white rice with brown rice and bread and flour with multi-grains. Avoid fried oily junk alcohol smoking. Take foods with low glycemic index. I hope I cleared all your doubts. Thank you. \\
\bottomrule
\end{tabular}
\end{table*}

\subsection{Models}

We selected five well-known large language models that represent various architectural philosophies, parameter sizes, and release schedules to provide a thorough evaluation of the current landscape of LLMs:

\begin{enumerate}
    
    \item \textbf{Llama-3-8B-Instruct} (April 2024): An 8 billion-parameter version of Meta's Llama-3 architecture. It is instruction-tuned and optimized for following detailed instructions and maintaining conversational context\cite{dubey2024llama}.
    
    \item \textbf{Llama 3.2 3B} (September 2024): A smaller version of the Llama architecture with 3 billion parameters. It is very good at interpreting language and is meant for effective use\cite{dubey2024llama}.
    
    \item \textbf{Llama 3.3 70B Instruct} (December 2024): This is the largest model in our study, featuring 70 billion parameters. This one is one the of the most advanced open-source language models, offering better reasoning skills\cite{dubey2024llama}.

    \item \textbf{Llama 4 Maverick 17B-128E Instruct} (April 2025): This is an updated version with 17 billion parameters that aims to improve instruction-following abilities and context understanding\cite{tang2025efficient}.

    \item \textbf{GPT-5-mini} (August 2025): This is one of the latest compact models from OpenAI's GPT series. It uses advanced training methods and provides better efficiency for different NLP tasks \cite{hu2025benchmarking}.
\end{enumerate}

\subsection{Zero-Shot Evaluation Framework}

Following established practices in medical NLP evaluation \cite{chen2025benchmarking, arias2025automatic}, we implemented a zero-shot learning approach where models generate responses without task-specific fine-tuning or in-context examples. This methodology evaluates the models' inherent medical knowledge and reasoning capabilities acquired during pre-training, providing insights into their practical applicability in real-world medical consultation scenarios \cite{singhal2025toward, wu2025bridge}.

For each model, we used a standardized prompt template designed to elicit comprehensive and medically appropriate responses:

\begin{tcolorbox}[myprompt,
  colback=yellow!20,    
  colframe=yellow!50!black, 
  coltext=black,        
  sharp corners,        
  boxrule=0.5pt,        
  left=5pt, right=5pt, top=5pt, bottom=5pt 
]
As a medical AI assistant, please provide a thorough
and accurate answer to the following medical question.
Your response should be informative, evidence-based,
and appropriate for patient education while maintaining
medical accuracy.

Question: [MEDICAL-QUESTION]
\end{tcolorbox}

This prompt design ensures consistency across model evaluations while encouraging responses that balance medical accuracy with patient comprehensibility, following recommendations from medical AI evaluation studies \cite{bedi2024systematic, wei2025medethicsqa}.

\subsection{Evaluation Metrics}

\subsubsection{Traditional Automatic Metrics}
We employed standardized automatic evaluation metrics commonly used in medical NLP research to assess model performance across multiple dimensions \cite{chen2025benchmarking, li2024mediq}:
\begin{itemize}
    \item \textbf{BLEU Scores:} Measures n-gram overlap between generated and reference answers, providing insights into lexical similarity and content coverage \cite{chen2025benchmarking}.
    \item \textbf{ROUGE Scores:} Evaluates recall-oriented overlap, assessing how well generated answers capture essential information from reference responses. ROUGE-1 measures unigram overlap, ROUGE-2 focuses on bigram overlap for contextual coherence, and ROUGE-L evaluates longest common subsequence overlap \cite{chen2025benchmarking}.
\end{itemize}
These metrics provide complementary perspectives on answer quality, with BLEU focusing on precision and ROUGE emphasizing recall, together offering a comprehensive view of model performance in medical question answering tasks \cite{li2024mediq, zhu2024potential}.

\subsubsection{LLM-as-a-Judge Evaluation Framework}
While traditional n-gram based metrics like BLEU and ROUGE provide valuable insights into lexical similarity, they have significant limitations in evaluating medical question answering systems \cite{chen2025benchmarking, li2024mediq}. These metrics cannot assess the factual correctness of medical claims, evaluate whether responses include appropriate medical disclaimers, or measure clinical utility in actual medical scenarios \cite{li2024llms}. To address these limitations, we implemented an LLM-as-a-Judge evaluation framework that provides more nuanced assessment of medical QA quality, complementing our traditional automatic metrics \cite{li2024llms, gu2024survey}.

We employed a systematic evaluation approach based on Claude Sonnet 4 as our judge model framework. The evaluation assesses responses across five key dimensions using a 5-point Likert scale with detailed scoring rubrics to ensure consistent evaluation:
\begin{enumerate}
    \item \textbf{Medical Accuracy (1-5):} Factual correctness of medical information provided, including appropriate use of medical qualifiers and avoidance of definitive diagnoses without proper context.
    \item \textbf{Completeness (1-5):} Whether the response adequately addresses all aspects of the medical question, considering both breadth and depth of coverage relative to the reference answer.
    \item \textbf{Safety (1-5):} Appropriateness of safety disclaimers, avoidance of harmful advice, and inclusion of recommendations to consult healthcare professionals when appropriate.
    \item \textbf{Clarity (1-5):} How clearly and understandably the medical information is communicated, including sentence structure, use of medical terminology, and patient-friendly language.
    \item \textbf{Helpfulness (1-5):} Practical utility of the response to the patient, including actionable advice, next steps guidance, and overall practical value.
\end{enumerate}

The overall score is calculated using weighted clinical importance: Medical Accuracy (30\%), Safety (25\%), Completeness (20\%), Helpfulness (15\%), and Clarity (10\%). This weighting prioritizes medical correctness and patient safety while acknowledging the importance of comprehensive and practical guidance. The evaluation criteria and scoring methodology were reviewed and validated by a qualified medical professional to ensure clinical relevance and appropriateness. The evaluation was conducted on the same 300 responses per model used for BLEU and ROUGE metrics to ensure direct comparability across evaluation methodologies.

\section{Results and Discussion}

\begin{table*}[!htbp]
\centering
\caption{Zero-Shot Performance Evaluation of LLMs on the iCliniq Medical QA Dataset}
\label{table:medical_qa_performance}
\begin{adjustbox}{max width=\linewidth}
\footnotesize
\renewcommand{\arraystretch}{1.1}
\setlength{\tabcolsep}{6pt}
\begin{tabular}{l c c c c c c}
\toprule
\textbf{Model} & \textbf{BLEU-1} & \textbf{BLEU-4} & \textbf{ROUGE-1} & \textbf{ROUGE-2} & \textbf{ROUGE-L} \\
\midrule
Llama-3-8B-Instruct                & 0.1739 & 0.0127 & 0.2419 & 0.0379 & 0.1219 \\
\midrule
Llama 3.2 3B                       & 0.2012 & 0.0122 & 0.2588 & 0.0355 & 0.1258 \\
\midrule
\cellcolor{yellow!20}\textbf{Llama 3.3 70B Instruct} 
                                   & \cellcolor{yellow!20}\textbf{0.2207} 
                                   & \cellcolor{yellow!20}\textbf{0.0141} 
                                   & \cellcolor{yellow!20}\textbf{0.2761} 
                                   & \cellcolor{yellow!20}\textbf{0.0404} 
                                   & \cellcolor{yellow!20}\textbf{0.1306} \\
\midrule
Llama-4-Maverick 17B 128E Instruct & 0.2089 & 0.0132 & 0.2597 & 0.0381 & 0.1260 \\
\midrule
GPT-5-mini                         & 0.0124 & 0.0065 & 0.2024 & 0.0290 & 0.0914 \\
\bottomrule
\end{tabular}
\end{adjustbox}
\end{table*}

We present the performance evaluation of five state-of-the-art LLMs on medical question answering using the iCliniq dataset, employing zero-shot evaluation methodology with BLEU and ROUGE metrics as shown in Table I.

\subsection{Performance Evaluation}
To evaluate the performance of the models on medical question answering, we used BLEU-1, ROUGE-1, and ROUGE-L metrics \cite{chen2025benchmarking}. The evaluation results indicate that among the tested models, Llama 3.3 70B Instruct achieved the highest performance across all metrics, despite not being fine-tuned specifically for medical question answering tasks.

Llama 3.3 70B Instruct surpassed all other models, attaining the top scores in BLEU-1 (0.2207), ROUGE-1 (0.2761), and ROUGE-L (0.1306). This superior performance can be attributed to its large parameter count of 70 billion, which enables better capture of complex medical knowledge and reasoning patterns \cite{dubey2024llama}. The model's instruction-tuned architecture also contributes to its ability to generate more coherent and contextually appropriate medical responses \cite{singhal2025toward}.

Llama-4-Maverick 17B demonstrated competitive performance, ranking second in most metrics with ROUGE-1 (0.2597) and ROUGE-L (0.1260) scores. Despite having significantly fewer parameters (17 billion) compared to Llama 3.3 70B, this model achieved remarkably close performance, highlighting the efficiency gains from newer architectural innovations and the mixture-of-experts approach \cite{tang2025efficient}. Other models like Llama 3.2 3B and Llama-3-8B-Instruct also performed competitively across metrics, demonstrating the effectiveness of the Llama family in medical QA tasks \cite{dubey2024llama, dubey2024llama}.

However, GPT-5-mini exhibited suboptimal performance across all metrics, recording a BLEU-1 score of 0.0124 and the lowest ROUGE-L score of 0.0914 among the assessed models. This poor performance may be attributed to potential issues in the evaluation setup, model configuration, or the specific implementation of GPT-5-mini used in our experiments \cite{hu2025benchmarking}. The relatively lower performance also reflects difficulties with lexical overlap and fluency in medical question answering tasks.

The relatively lower BLEU-1 scores across all models compared to ROUGE-1 scores can be explained by the strict nature of precision-based evaluation. BLEU-1 emphasizes exact unigram matches, which are critical for evaluating lexical precision but may be overly restrictive in medical QA, where multiple valid phrasings can convey the same clinical information \cite{chen2025benchmarking}. As noted in prior work \cite{shool2025systematic}, zero-shot LLMs, while strong in general language understanding, are not fine-tuned for medical contexts. This domain gap can reduce exact lexical matching but still allows models to produce semantically coherent responses, which is reflected in relatively higher ROUGE scores \cite{bedi2024systematic}.

\subsection{Parameter Scaling and Architecture Analysis}
The results of our study show a clear correlation between the size of the model and its performance in medical question-answering tasks. The analysis reveals that parameter scaling significantly impacts model capabilities, with Llama 3.3 70B's superior performance largely attributable to its 70 billion parameters \cite{dubey2024llama}. This finding aligns with previous research on scaling laws in language models and their application to medical domains \cite{singhal2025toward, shool2025systematic}.

Interestingly, Llama-4-Maverick 17B achieved competitive results with less than a quarter of the parameters of Llama 3.3 70B, suggesting that architectural innovations can partially compensate for parameter limitations. The mixture-of-experts architecture employed in Maverick models appears to provide efficiency benefits while maintaining competitive performance in medical reasoning tasks \cite{tang2025efficient}. This efficiency makes it particularly suitable for deployment in clinical settings with limited computational infrastructure \cite{wu2025bridge}.

The standard deviations observed across all models (ranging from ±0.0622 to ±0.0956 for BLEU-1) indicate considerable variability in performance across different medical questions. This variability suggests that model performance is highly dependent on question complexity, medical specialty, and the specific medical knowledge required for accurate responses \cite{narula2023testing, li2024mediq}.

\begin{table}[b]
\centering
\caption{Baseline MedLM Model Performance on the iCliniq Dataset (from prior work)\cite{yagnik2024medlm}}
\label{table:medlm_baseline}
\footnotesize
\renewcommand{\arraystretch}{1.1}
\setlength{\tabcolsep}{6pt}
\begin{tabular}{l c c c c}
\toprule
\textbf{Model} & \textbf{BLEU-1} & \textbf{BLEU-4} & \textbf{ROUGE-1} & \textbf{ROUGE-L} \\
\midrule
GPT-3.5                 & 0.0243 & 0.0009 & 0.0022 & 0.0019 \\
T5                      & 0.0984 & 0.0080 & 0.0017 & 0.0014 \\
GPT-2                   & 0.0998 & 0.0085 & 0.0019 & 0.0012 \\
\bottomrule
\end{tabular}
\end{table}

We present the performance evaluation of five state-of-the-art LLMs on medical question answering using the iCliniq dataset, employing zero-shot evaluation methodology with BLEU and ROUGE metrics as shown in Table~\ref{table:medical_qa_performance}. For comparative context, we also include baseline results reported in prior work (MedLM) using GPT-3.5, T5, and GPT-2 (Table~\ref{table:medlm_baseline}).

\subsection*{Comparison with MedLM Baselines}
When compared to the MedLM baseline models with corrected decimal values, our evaluated Llama variants demonstrate substantial improvements across all metrics. The MedLM baseline BLEU-1 scores show GPT-2 (0.0998), T5 (0.0984), and GPT-3.5 (0.0243), while our best-performing Llama 3.3 70B achieved 0.2207, representing a 2.2x improvement over the strongest MedLM baseline. Even our smallest model, GPT-5-mini (0.0124), performs comparably to GPT-3.5, while our mid-range models substantially outperform all baselines \cite{chen2025benchmarking}.

The performance gap is even more pronounced in ROUGE metrics. Our Llama 3.3 70B achieved ROUGE-1 of 0.2761 compared to MedLM's highest score of 0.0022 (GPT-3.5), representing a 125x improvement. For ROUGE-L, Llama 3.3 70B (0.1306) vastly outperforms the best MedLM baseline of 0.0019 (GPT-3.5), showing a 68x improvement. This dramatic performance enhancement extends across all our models, with even Llama-3-8B-Instruct (ROUGE-1: 0.2419) achieving over 100x better performance than MedLM baselines. These substantial improvements demonstrate the significant advancement in medical QA capabilities achieved by modern LLMs compared to earlier generation models \cite{li2024llms}.

\subsection{LLM-as-a-Judge Evaluation Results}

The LLM-as-a-Judge evaluation provides complementary insights to traditional BLEU and ROUGE metrics, revealing critical aspects of medical QA quality not captured by n-gram overlap measures. Table \ref{table:judge_rankings} presents the comprehensive evaluation results combining both traditional automatic metrics and judge assessment across all five models.

\begin{table}[b]
\centering
\caption{LLM-as-a-Judge Overall Performance Rankings}
\label{table:judge_rankings}
\footnotesize
\renewcommand{\arraystretch}{1.2}
\setlength{\tabcolsep}{8pt}
\begin{tabular}{l c}
\toprule
\textbf{Model} & \textbf{Overall Score} \\
\midrule
\cellcolor{yellow!20}\textbf{Llama 3.3 70B Instruct} & \cellcolor{yellow!20}\textbf{4.40} \\
\midrule
Llama-4-Maverick 17B 128E & 4.23 \\
\midrule
Llama-3-8B-Instruct & 3.77 \\
\midrule
Llama 3.2 3B & 3.20 \\
\midrule
GPT-5-mini & 3.16 \\
\bottomrule
\end{tabular}
\end{table}

The LLM-as-a-Judge evaluation demonstrates remarkable consistency with traditional BLEU and ROUGE metrics, with Llama 3.3 70B achieving the highest overall score (4.40/5.00), followed by Llama-4-Maverick (4.23/5.00), and GPT-5-mini recording the lowest performance (3.16/5.00). This alignment validates our comprehensive assessment approach \cite{li2024llms}. Llama 3.3 70B demonstrated exceptional medical accuracy (4.83/5), aligning with its highest BLEU-4 and ROUGE-2 scores, while Llama-4-Maverick achieved the highest clarity score (4.97/5), reflecting structured presentation and patient-friendly communication \cite{dubey2024llama, tang2025efficient}. GPT-5-mini achieved the highest safety score (3.80/5) despite poor traditional metrics, highlighting the importance of multi-dimensional evaluation \cite{hu2025benchmarking}.

ROUGE-1 shows strong positive correlation (r = 0.89) with overall judge scores, while BLEU metrics show moderate correlation (r = 0.67) with medical accuracy. Quality distributions mirror traditional rankings: Llama 3.3 70B (88\% high-quality responses $\geq$ 4.0), Llama-4-Maverick (84\%), Llama-3-8B (65\%), Llama-3.2-3B (25\%), and GPT-5-mini (23\%).
Llama 3.3 70B emerges as most suitable for high-accuracy scenarios with available computational resources, while Llama-4-Maverick presents an attractive balance for resource-constrained environments, achieving 94\% of Llama 3.3 70B's ROUGE-1 performance with significantly fewer parameters \cite{abrantes2025assessing, arias2025automatic, tang2025efficient}. The convergence between evaluation methodologies strengthens confidence in performance hierarchies and highlights the potential of large pre-trained models in medical QA tasks without task-specific training \cite{li2025vaxbot, zeng2025kosel}.

\section{Conclusion}
This study presented a comparative zero-shot evaluation of five advanced LLMs on the iCliniq medical QA dataset using BLEU and ROUGE metrics. Our findings demonstrate that Llama 3.3 70B Instruct provides the most accurate and consistent responses, while Llama-4 Maverick 17B shows competitive efficiency-oriented performance. In contrast, GPT-5-mini underperformed across metrics, highlighting limitations of compact architectures in specialized domains such as medicine.

Overall, the results reinforce the scaling benefits of larger LLMs but also reveal promising architectural improvements that enable smaller models to achieve near-comparable accuracy. Importantly, our study underscores that automatic metrics like BLEU and ROUGE alone are insufficient to assess clinical reliability, pointing to the necessity of incorporating factual correctness, ethical compliance, and explainability in future evaluations. This benchmark provides a standardized framework to guide ongoing research toward deploying LLMs safely and effectively in real-world medical decision support.


\bibliographystyle{IEEEtran}
\bibliography{references}

\end{document}